\documentclass{vgtc}                          

\ifpdf
  \pdfoutput=1\relax                   
  \usepackage{graphicx}                
  \DeclareGraphicsExtensions{.pdf,.png,.jpg,.jpeg} 
\else
  \usepackage{graphicx}                
  \DeclareGraphicsExtensions{.eps}     
\fi%

\graphicspath{{figures/}{pictures/}{images/}{./}} 

\usepackage{microtype}                 
\PassOptionsToPackage{warn}{textcomp}  
\usepackage{textcomp}                  
\usepackage{mathptmx}                  
\usepackage{times}                     
\usepackage{cite}                      
\usepackage{tabu}                      
\usepackage{booktabs}                  

\onlineid{0}

\vgtccategory{Research}

\vgtcinsertpkg

\title{Diffusion Attack: Leveraging Stable Diffusion for\\
Naturalistic Image Attacking}

\author{Qianyu Guo\thanks{e-mail: guo716@purdue.edu}\\ %
        \scriptsize  Purdue University %
\and Jiaming Fu\thanks{e-mail: fu330@purdue.edu}\\ %
     \scriptsize Purdue University %
     \and Yawen Lu\thanks{e-mail: lu976@purdue.edu}\\ %
     \scriptsize Purdue University %
\and Dongming Gan\thanks{e-mail: dgan@purdue.edu}\\ %
     \parbox{1.4in}{\scriptsize \centering  Purdue University}}

\abstract{In Virtual Reality (VR), adversarial attack remains a significant security threat. Most deep learning-based methods for physical and digital adversarial attacks focus on enhancing attack performance by crafting adversarial examples that contain large printable distortions that are easy for human observers to identify. However, attackers rarely impose limitations on the naturalness and comfort of the appearance of the generated attack image, resulting in a noticeable and unnatural attack. To address this challenge, we propose a framework to incorporate style transfer to craft adversarial inputs of natural styles that exhibit minimal detectability and maximum natural appearance, while maintaining superior attack capabilities. 
} 

\CCScatlist{
  \CCScatTwelve{Computing methodologies}{Artificial intelligence}{Computer vision}{};
  \CCScatTwelve{Computing methodologies}{Computer graphics}{Image manipulation}{Image processing}
}

\begin{document}

\firstsection{Introduction}

\maketitle

While deep neural networks (DNNs) have proven their effectiveness in a wide range of applications, delivering impressive performance in areas such as object detection \cite{lu2019geometric}, autonomous driving, and face recognition, they can also pose a threat to the security of a wide range of applications by compromising the security of applications or collecting private information from the public. Especially in Virtual Reality systems, adversarial attacks are able to manipulate or exploit weaknesses to deceive or compromise its functionality.

Neural style transfer is a novel technique that uses Convolutional Neural Networks (CNNs) to transform content images into different styles. In this work, we capitalize on previous advances in neural style transfer and focus on proposing a novel technique to effectively and successfully interfere with adversarial examples while maintaining the most natural and deceptive appearance on the target objects. With a latent text-to-image diffusion model capable of generating photorealistic images, we are able to easily provide a list of original text prompts in the generation of hundreds of style images as adversarial attack examples, which allows full control of the patterns rather than relying on the limited input reference style image. Thus, our goal for style generation is to produce a number of object styles (e.g., patterns on clothes, spray paint on cars) without significantly changing their inherent shapes and appearance. 

Our research centres on generating the most natural-looking and undetectable style-transformed adversarial samples and framing them within a specially designed adversarial image attack network. In this work, we use a pre-trained Deep Stable Diffusion model to generate various style-transformed images for targeted image attack by providing original text prompts, then conduct a thorough qualitative analysis of the proposed method to verify that the generated adversarial attack images are of the most natural and deceptive appearance, while still maintaining superior attack performance. Furthermore, we employ various non-reference image quality assessment (NR-IQA) and image aesthetic assessment models to evaluate the perceptual quality of the generated adversarial images for the first time, further confirming that our approach can produce natural and visually high-quality images in a quantitative manner \cite{guo2021multi}.

\begin{figure}[tb]
 \centering 
 \includegraphics[width=\columnwidth, height=5.4cm]{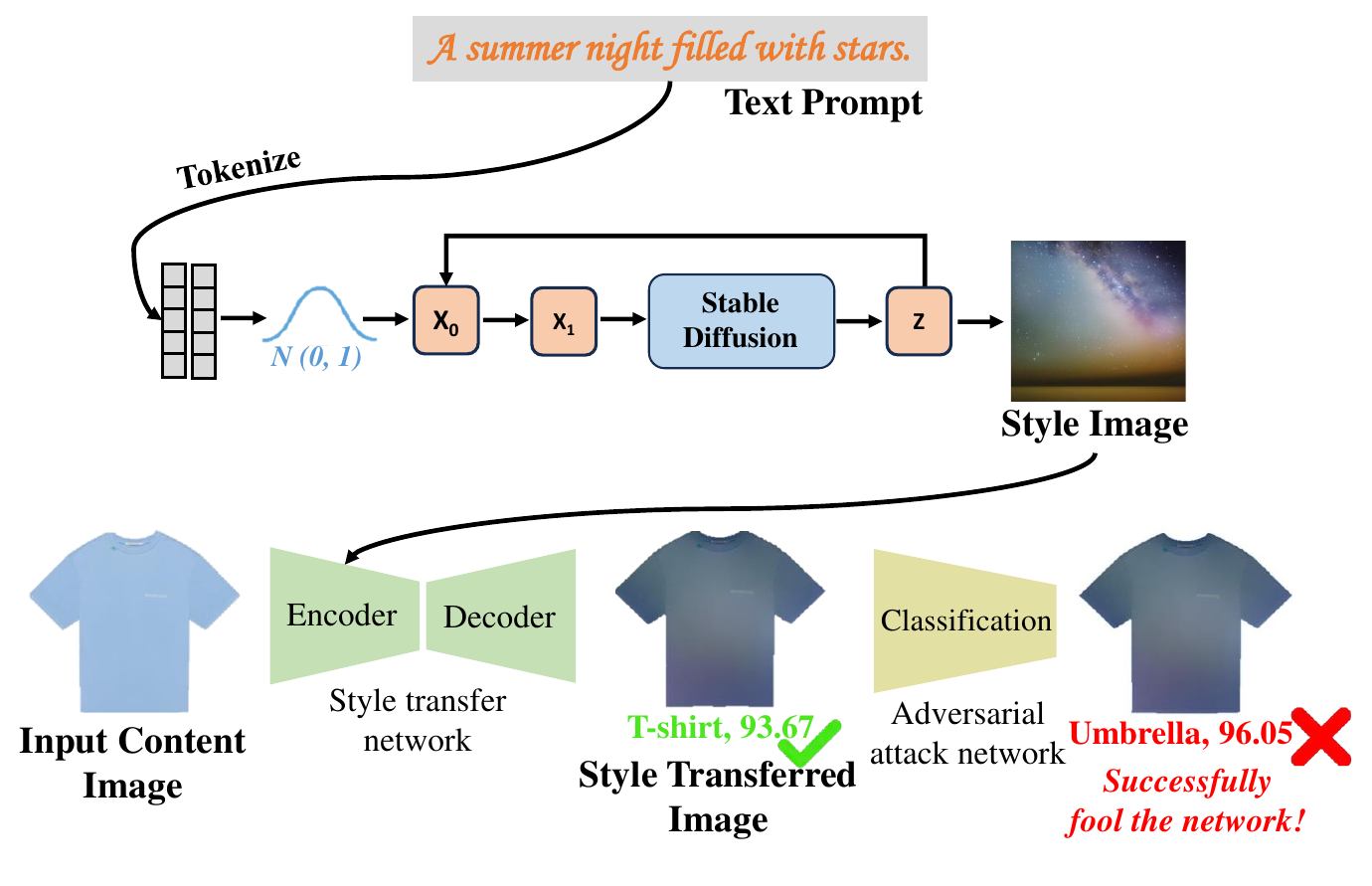}
 \vspace{-7mm}
 \caption{{Overview of the proposed \textit{\textbf{Diffusion Attack}}.} Starting from the text prompt to generate a style image, our adversarial attack model causes the style transferred image to be classified as a fake umbrella instead of the original T-shirt. This is in contrast to existing attack methods that invade the system with noticeable and unnatural noises.}
 \label{fig:sample}
 \vspace{-2mm}
\end{figure}

\section{Diffusion Attack}

As a latent diffusion model, Stable Diffusion is designed to create detailed images based on provided textual prompts. The model consists of three components: encoder network, U-Net, and decoder network. The encoder network is used to convert input text into a latent representation. Subsequently, the U-Net progressively diffuses information to generate novel data. The final image is then created by utilizing this processed data while maintaining a high degree of semantic coherence between the original prompt and the output.

As shown in Figure \ref{fig:sample}, based on the obtained style image from the Stable Diffusion, we implement neural style transfer to impart a new naturalistic style to the content image. To enable the network to concentrate on the area of interest, we incorporate a mask image as well. As the deeper layer of a pre-trained convolutional network transforms the input image into feature maps that prioritize content over texture or color pixel details, the content loss $l_{content}$ is determined by computing the Mean Square Error (MSE) between the feature maps of the original content image and the generated image. To compute the style loss, features are extracted from multiple intermediate layers and the feature correlations are given by the Gram matrix $Gram$. The goal of using style loss is to minimize the gap in correlation scores between the style feature maps and the input feature maps, which means that style correlations should be as consistent as possible. The style loss $l_{style}$ is also obtained by computing MSE between the Gram matrix-based style representations of the style image $Gram_{sl}$ and generated image $Gram_{xl}$. 

After obtaining generated style image, we build up an image attack neural network to generate an attacked stylized image. This process would continually attack the existing classifier goal-oriented, and force it to predict a wrong class label. To fool the pre-trained classifier Inception V3, we use the prevalent cross-entropy loss to define the adversarial loss $l_{adv}$. To decrease the variation between adjacent pixels, we consider using the smoothness loss $l_{smooth}$ to promote the adversarial images with low variance local patches. This enhances the naturalness and smoothness of generated images.

The overall loss function $l_{total}$ of the proposed methodology can be summarized as:

\begin{equation}
l_{total} = \lambda_{1} l_{content} + \lambda_{2} l_{style} + \lambda_{3} l_{adv} + \lambda_{4} l_{smooth}
\end{equation}

\begin{figure}[t!]
\begin{center}
\includegraphics[width=0.97\linewidth,height=50mm]{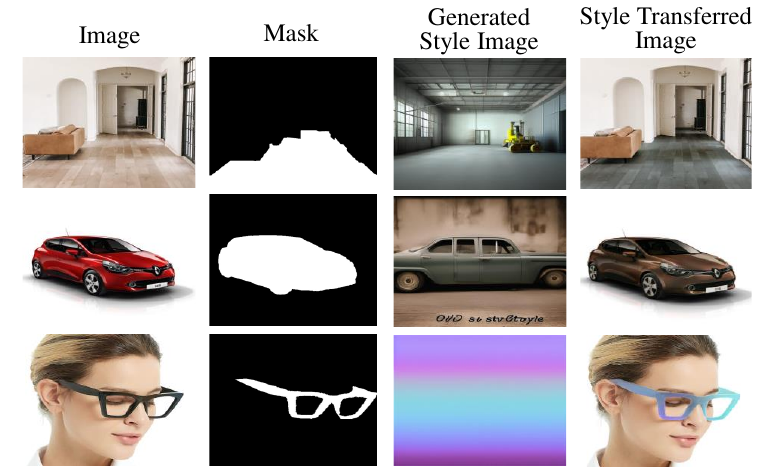} 
\end{center}
\vspace{-5mm}
   \caption{Style transferred image results from different provided style images. Our \textit{Diffusion Attack} has advantages in rendering realistic textures and coherent structures on the target object and region.
  }
  \vspace{-3mm}
\label{style}
\end{figure}

\section{Experiment and Preliminary Results}

In this poster, we build the entire algorithm in a two-stage training using our own style image data. To prepare the data for training, we resize each input image to dimensions of 299 $\times$ 299. We then perform a normalization to ensure that the pixel values fall within the range of -1 to 1, as required by the Inception-v3 model used. Our style transfer training sub-stage uses the L-BFGS optimizer with a learning rate set to 1.0 and a batch size of 4. For the adversarial attack sub-stage, we build on a pre-trained Inception-v3 model for feature extraction and aggregation with a modified auxiliary classifier to propagate target label information to feed the classification network. 

We evaluate our generated style transfer images in Figure \ref{style} on images from ImageNet and widely collected data from the Internet. It can be observed that for different scenes (e.g., floor, car, and glasses), our style transfer network performs realistic and naturalistic generation with the prompt-driven style images. The newly generated objects maintain the original object boundaries, geometry and shapes, while appearing in different materials, colors, and textures.

In addition to the qualitative comparisons, we also perform a set of quantitative evaluations to assess the image aesthetic and quality of the images generated by our approach compared to other attack baselines and real content images. NIMA\cite{talebi2018nima} and Topiq\_iaa \cite{chen2023topiq} are image aesthetic assessment models, which are able to account for the subjective factors by taking into consideration the diversity of human preferences. Topiq\_nr \cite{chen2023topiq} and Tres \cite{golestaneh2022no} evaluate the perceptual quality of the images, exploiting multi-scale features without neglecting the importance of semantic guidance. As shown in Table \ref{table}, our proposed methods achieve significantly higher aesthetic and quality scores than other baselines on four standard metrics. 

\begin{table}[]
\resizebox{\linewidth}{!}{
\renewcommand\arraystretch{1.35}
\begin{tabular}{c|c|c|c|c}
                                 & \begin{tabular}[c]{@{}c@{}}NIMA $\uparrow$ \\ (0$\sim$10)\end{tabular} & \begin{tabular}[c]{@{}c@{}}Topiq\_iaa $\uparrow$ \\ (0$\sim$10)\end{tabular} & \begin{tabular}[c]{@{}c@{}}Topiq\_nr $\uparrow$\\ (0$\sim$1)\end{tabular} & \begin{tabular}[c]{@{}c@{}}Tres $\uparrow$\\ (0$\sim$100)\end{tabular} \\ \hline
Content image                    & 5.37                                                   & 4.92                                                         & 0.70                                                      & 82.56                                                  \\ \hline
\textbf{Diffusion Attack (Ours)} & \textbf{4.78}                                          & \textbf{4.46}                                                & \textbf{0.62}                                             & \textbf{72.32}                                         \\ \hline
Woitschek et al. \cite{woitschek2021physical}                 & 2.97                                                   & 3.29                                                         & 0.28                                                      & 22.21                                                  \\ \hline
SLAPs \cite{lovisotto2021slap}                            & 3.73                                                   & 3.83                                                         & 0.48                                                      & 49.18                                                  \\ \hline
\end{tabular}}
\vspace{0.5mm}
\caption{Our \textit{Diffusion Attack} is able to achieve higher image quality and aesthetic assessment average scores compared with baselines. Higher scores indicate better performance.}
\label{table}
\vspace{-2mm}
\end{table}

\begin{figure}[t!]
\begin{center}
\includegraphics[width=0.97\linewidth,height=45mm]{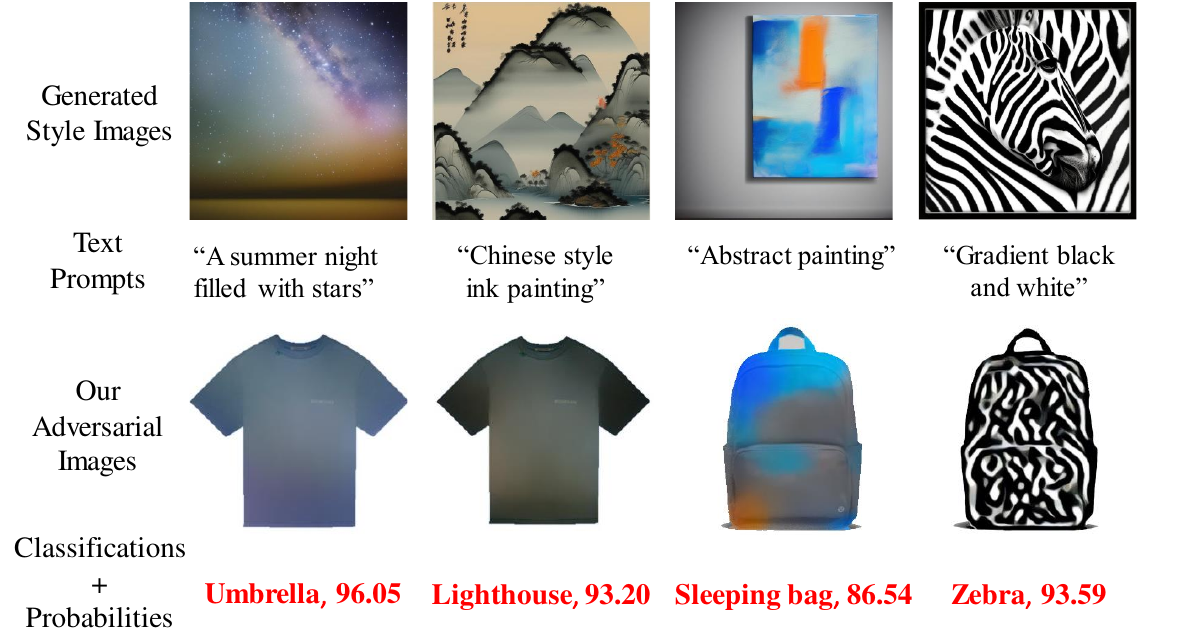} 
\end{center}
\vspace{-5mm}
   \caption{Attack performance shows the expected misclassification label and its probability. \textit{Diffusion Attack} can generate images with various adversarial patterns combining different textures and colors. 
  }
\label{attack}
\vspace{-3mm}
\end{figure}

We perform adversarial attack evaluations on several common everyday objects (e.g., backpacks and T-shirts). As shown in Figure \ref{attack}, our adversarial T-shirt can be misclassified as "umbrella" and "lighthouse" based on different provided text promotions, where the confidence of the classification network is high and both over 93\%. Similarly, backpacks can be misclassified as "sleeping bag" and "zebra" with an average confidence of around 90\%. 

\section{Conclusion}

In this poster, we propose a new naturalistic diffusion-based adversarial image attack approach, named \textit{\textbf{Diffusion Attack}}, by exploiting the joint training of style transfer network and adversarial attack network. With the capability of attacking image style transfer and generation, our approach successfully generates naturalistic adversarial images while maintaining the competitive attacking performance, based on the extensive experiment on qualitative and quantitative with other recent image attacking methods. Furthermore, we provide a novel non-reference perceptual image quality assessment method to provide a numerical comparison of the quality of the generated samples beyond the widely used visual comparison alone. We intend to conduct further analysis of the framework's impact in the future.


\bibliographystyle{abbrv-doi}

\bibliography{template}
\end{document}